\documentclass[10pt,twocolumn,letterpaper]{article}

\usepackage{cvpr}              % To produce the CAMERA-READY version
\definecolor{cvprblue}{rgb}{0.21,0.49,0.74}
\usepackage[pagebackref,breaklinks,colorlinks,allcolors=cvprblue]{hyperref}
\usepackage{multirow}
\usepackage{algorithmic}
\usepackage{algorithm}
\usepackage{graphicx} % Required for inserting images
\usepackage{amsmath} 
\usepackage{algorithm} 
\usepackage{algorithmic} 
\usepackage{array}
\usepackage[accsupp]{axessibility}
\def\paperID{13624} % *** Enter the Paper ID here
\def\confName{CVPR}
\def\confYear{2025}

\title{Deterministic Image-to-Image Translation via Denoising Brownian Bridge Models with Dual Approximators}

\author{
   \textbf{ Bohan Xiao$^{1}$\thanks{Both authors contributed equally to this research.},  
    Peiyong Wang$^{2}$\footnotemark[1],  
    Qisheng He$^{1}$,  
    Ming Dong$^{1}$ }\\
    Department of Computer Science, Wayne State University $^{1}$, Detroit, Michigan, USA\\
    Department of Mathematics, Wayne State University$^{2}$, Detroit, Michigan, USA\\
    {\tt\small \{gk6511, pywang, qisheng.he, mdong\}@wayne.edu}
}

\begin{document}

\maketitle

\begin{abstract}
Image-to-Image (I2I) translation involves converting an image from one domain to another. Deterministic I2I translation, such as in image super-resolution, extends this concept by guaranteeing that each input generates a consistent and predictable output, closely matching the ground truth (GT) with high fidelity. In this paper, we propose a denoising Brownian bridge model with dual approximators (Dual-approx Bridge), a novel generative model that exploits the Brownian bridge dynamics and two neural network-based approximators (one for forward and one for reverse process) to produce faithful output with negligible variance and high image quality in I2I translations. Our extensive experiments on benchmark datasets including image generation and super-resolution demonstrate the consistent and superior performance of Dual-approx Bridge in terms of image quality and faithfulness to GT when compared to both stochastic and deterministic baselines. Project page and code: https://github.com/bohan95/dual-app-bridge
\end{abstract}    
\section{Introduction}
\label{sec:intro}
%-------------------------------------------------------------------------
Image-to-Image (I2I) translation is a significant area of research with broad applications across academia and industry. It involves transforming an image from one domain to another while preserving essential features. Key tasks in this field include generating realistic images from semantic labels \cite{pix2pix, cycleGAN, c_sde, LDM}, transforming segmentation masks into photo-like outputs \cite{DDBM, LDM, A_bridge}, and enhancing image quality through super-resolution \cite{SRDiff, chen2023solving, RankSRGAN}. 

Among the wide range of applications, some I2I tasks particularly benefit from deterministic models. For instance, in super-resolution, the goal is to translate a low-resolution image into a high-resolution one while accurately preserving fine details. This task, as shown in recent studies \cite{wu2024diffusion, tsao2024boosting}, requires consistency and high faithfulness to ground true (GT) because for a given low-resolution image, there is typically one ``correct'' high-resolution output that should be produced. In such an I2I task, we note that \textit{faithfulness or fidelity is to the individual GT}, not just to the distribution of GT. Any variability in the generated output could result in artifacts or loss of critical information, making stochastic approaches less desirable. Similarly, in tasks such as denoising \cite{Denoising_1, Denoising_2} or inpainting \cite{Inpainting_1, Inpainting_2}, where missing or noisy parts of an image need to be restored, a deterministic approach is preferred.

Generative Adversarial Networks (GANs) \cite{goodfellow2014gan} have been instrumental in advancing I2I translation, with models like Pix2Pix \cite{pix2pix} and CycleGAN\cite{cycleGAN} demonstrating impressive results. GANs are deterministic during inference, producing the same output for a given input, which is advantageous for tasks requiring faithfulness. However, GANs often struggle to generate high-quality images, frequently resulting in blurriness and artifacts \cite{sym12101705, s22218540}.

\begin{figure}
    \centering
    \includegraphics[width=0.5\textwidth]{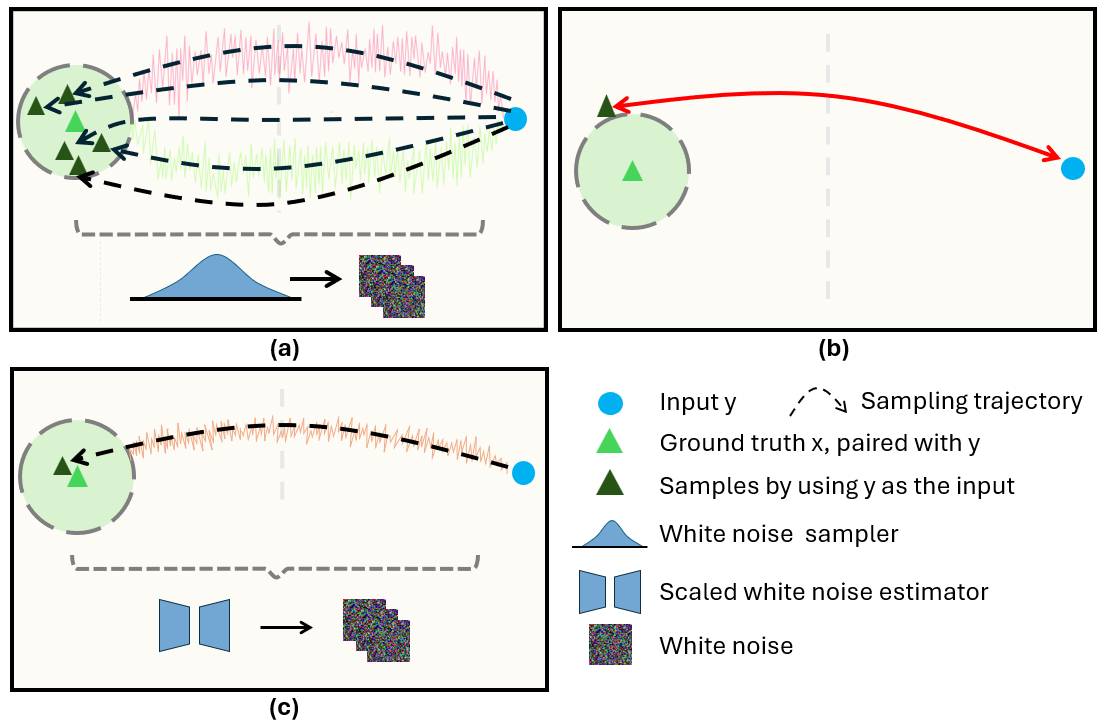}
    \caption{Sampling comparison between SDE-based sampler, PF-ODE-based sampler, and Dual-approx Bridge. The circular area within the dashed circle indicates where the SDE-based sampling outputs land. \textbf{(a)} SDE-based Sampling: diverse outputs of high image quality. \textbf{(b)} PF-ODE-based Sampling: deterministic outputs lack of fine details of ground truth (GT). \textbf{(c)} Dual-approx Bridge Sampling: deterministic output with fine details of GT.}
    \label{fig:structure}
    \vspace{-0.2in}
\end{figure}

Diffusion models, such as Denoising Diffusion Probabilistic Models (DDPM) \cite{DDPM} and Latent Diffusion Models (LDM) \cite{LDM}, generate high-quality images by progressively adding noise to images and learning the reverse process. 
Bridge models, such as Brownian bridge diffusion models \cite{BBDM, A_bridge}, utilize Stochastic Differential Equations (SDEs) to directly interpolate between input and output distributions, allowing precise control over transformations \cite{LDM, c_sde}. This approach improves analytical tractability, making bridge models well suited for I2I translations. 
However, the inherent stochasticity in these models (see Figure \ref{fig:structure}(a)) complicates tasks where faithful results are required (e.g., image super-resolution and image reconstruction). Unlike traditional diffusion models \cite{DDPM, LDM, sde}, which start from noise and gradually denoise the input to generate an output, bridge models start from structured data and interpolate between input and output distributions. This difference means that deterministic methods like Probability Flow Ordinary Differential Equations (PF-ODEs) \cite{sde}, when applied to bridge models, can lead to blurry images \cite{sde, DDBM}, resulting in degraded model performance (see Figure \ref{fig:structure}(b)). Therefore, there is a growing need for deterministic models that can preserve both image quality and faithfulness in I2I translations. 

To resolve this dilemma in the pursuit of deterministic sampling with both high image quality and fidelity, in this paper we propose a denoising Brownian bridge model with Dual Approximators (abbreviated as ``Dual-approx Bridge''), a generative model that exploits the Brownian bridge structure to produce faithful output with negligible variance and with an image quality better than or comparable to that generated by SDE-based samplers. More specifically, in the SDE-based structure of a Brownian bridge with both forward and corresponding reverse Brownian motions, we train a neural network in the forward diffusive process to estimate the score function and the initial state in the diffusive iteration. Meanwhile, in the reverse process of the Brownian bridge, a second neural network is trained to approximate the white noise scaled to the time step increment. Here, we emphasize that in deterministic I2I, our goal is to \textit{generate a unique output from a given input, rather than translating probability distributions}, as shown in Figure \ref{fig:structure}(c). The major contributions of this work are as follows:
\begin{itemize}
    \item To our best knowledge, Dual-approx Bridge is the first generative model that exploits the Brownian bridge structure to produce faithful output with negligible variance and high image quality for deterministic I2I translations. This is achieved through a novel design of a pair of approximators, trained in the forward and reverse diffusive processes, respectively. 

    \item In Dual-approx Bridge, the reverse approximator optimizes the reverse denoising process and presents an optimal deterministic route for the generative process. From a theoretical perspective, we analyze our model in a general setting where the white noise may take any time-dependent variance.
    
    \item Through extensive experiments on benchmark datasets including image generation and super-resolution, we demonstrate the consistent and superior performance of Dual-approx Bridge on image quality and faithfulness to GT when compared to state-of-the-art (SOTA) GAN-based and diffusion-based I2I models.
\end{itemize}

\section{Related Work}
\label{sec:related_work}

\begin{figure*}
    \centering
    \includegraphics[width=0.95\textwidth]{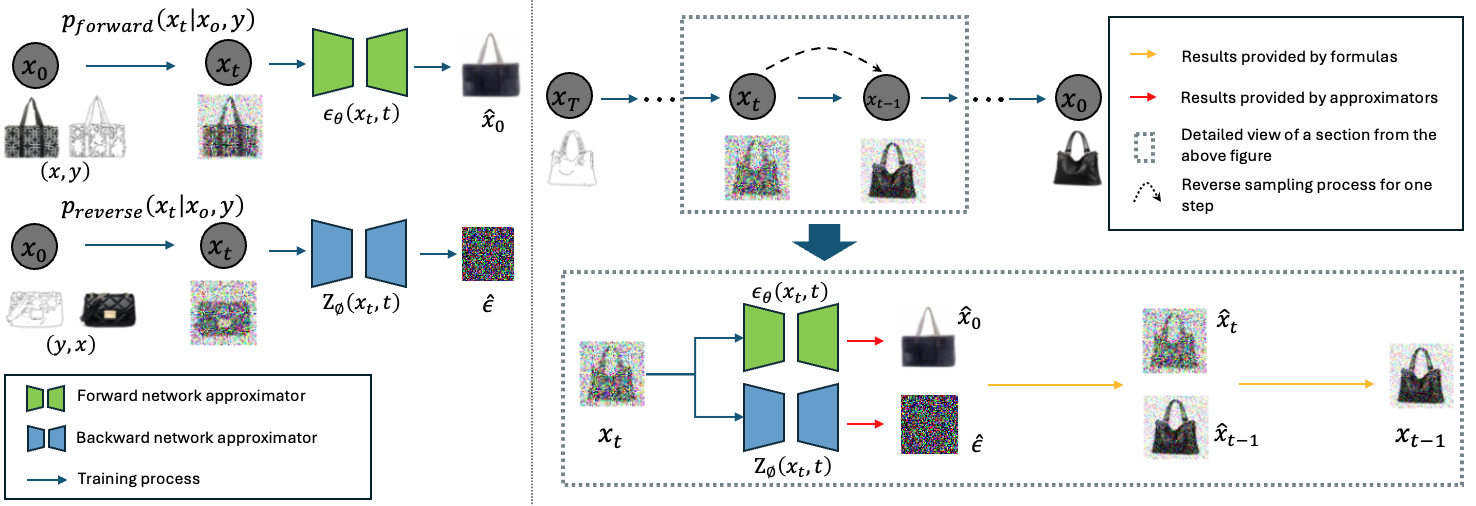}
    \caption{Training process of the two approximators (top-left: forward diffusive process, bottom-left: reverse diffusive process). Right: Sampling workflow with both approximators in action.}
    \label{fig:architecture}
    \vspace{-0.2in}
\end{figure*}

In this section, we briefly review the related literature, including image-to-image translation, GAN Models, Diffusion Models, and Bridge Models.

%-------------------------------------------------------------------------
\subsection{Image-to-Image Translation}

Image-to-image (I2I) translation has become a core task in computer vision \cite{pang2021image, sym12101705, s22218540}, focused on learning mappings between the source and target image domains for applications such as style transfer \cite{gatys2015neural} and colorization \cite{cheng2015deep}. Traditional methods, including Pix2Pix \cite{pix2pix} and CycleGAN\cite{cycleGAN}, utilize GAN-based architectures to translate images and support diverse I2I tasks by capturing domain-specific features, though challenges remain in maintaining high quality across complex domains \cite{borji2019pros, shmelkov2018good}. Deterministic I2I translation builds on this by ensuring that each input produces a consistent, predictable output — an essential requirement in fields like medical imaging \cite{litjens2017survey}, where precision and reproducibility are crucial. In tasks such as super-resolution \cite{dong2015image} or denoising \cite{ulyanov2018deep}, deterministic models can enhance image resolution and clarity without introducing variations that could adversely affect analysis or interpretation. %This predictability benefits applications requiring high faithfulness \cite{DDRM, SRDiff, ESRGAN}, where fidelity to the GT is of paramount importance. 
\subsection{Brown Bridge}
The Brownian Bridge (BB) \cite{A_bridge, BBDM} is a fundamental stochastic process that models a constrained Brownian motion, where the trajectory is conditioned on fixed start and end points. Unlike an unconstrained Wiener process, which evolves freely, a Brownian bridge interpolates between two known states while maintaining stochastic variations. This property makes it particularly useful in probabilistic modeling, including image-to-image (I2I) translation, conditional generative modeling, and stochastic optimal transport.
\subsection{Generative Adversarial Networks}
GANs \cite{goodfellow2014gan} have been pivotal in advancing I2I translation by learning mappings between image domains through adversarial training. One foundational model, Pix2Pix \cite{pix2pix}, introduced a conditional GAN approach for paired I2I tasks, enabling applications like colorization\cite{cao2017unsupervised} and semantic label-to-photo translation\cite{pix2pix}. For unpaired I2I tasks, CycleGAN \cite{cycleGAN} employs cycle consistency, allowing the model to translate images across domains without requiring paired samples.  However, GANs often face artifacts in generated images \cite{shmelkov2018good, borji2019pros, wang2021gan}. Such limitations in image quality, combined with the common challenges of training instability, make GANs less suitable for applications demanding both quality and faithfulness.

% Other GAN-based models, like the more recent MUNIT \cite{huang2018multimodal} and DRIT \cite{lee2018diverse}, further expand on I2I translation by enabling diverse and multimodal outputs.

%which recent work aims to address through improved architectures and loss functions \cite{wang2021gan}. Despite their ability to produce deterministic results, GAN models often struggle with generating high-quality images \cite{shmelkov2018good, borji2019pros}, with outputs that can appear unrealistic or suffer from blurriness. These issues are particularly evident in complex tasks where high image quality is essential. Such limitations in image quality, combined with the common challenges of training instability and potential artifacts, make GANs less suitable for applications demanding both quality and faithfulness.

\subsection{Diffusion and Bridge Models}
Diffusion models have established themselves as a leading approach in generative modeling \cite{dhariwal2021diffusion} due to their ability to produce high-quality, realistic images through a gradual denoising process. DDPM \cite{ho2020denoising} pioneered this approach, modeling image generation as a series of iterative steps in which random noise is gradually refined into a structured target image. The process results in exceptionally high-quality images with reduced artifacts, often surpassing GANs. LDMs \cite{LDM} further refined DDPM by focusing on a latent space rather than a pixel space, making the generation process more efficient. % while maintaining high image quality.

To model the denoising process, diffusion models can rely on two distinct mathematical frameworks: SDEs and PF-ODEs \cite{sde}. SDEs, characterized by their inherent randomness, allow the model to generate diverse outputs by gradually reversing noise, enhancing detail and realism, but lacking consistency. This randomness can lead to the same input producing varying outputs. PF-ODEs, on the other hand, remove this randomness, providing a deterministic pathway that produces consistent outputs at the expense of fine details and image richness \cite{BBDM, sde}. Balancing the high image quality provided by SDEs with the consistency achieved by PF-ODEs remains a core challenge in diffusion models.

Bridge models \cite{DDBM, UNSB, liu20232} integrate principles from both diffusion and bridge frameworks to create a probabilistic connection between input and output distributions. Both the Schrödinger and Brownian Bridge frameworks \cite{DDBM, A_bridge} have demonstrated significant improvements in image quality across various I2I tasks when compared to traditional diffusion models. However, they inherit certain limitations from both SDE-based and PF-ODE-based sampling. Specifically, they retain the stochastic characteristics of SDEs while also experiencing the lower image quality associated with PF-ODE-based samplers \cite{DDBM}.

%Leveraging the Schrödinger Bridge framework and incorporating EDM-inspired \cite{EDM} sampling techniques, \cite{DDBM} combines the strength of diffusion models in high-quality image synthesis with the stability and structured transformation pathways of bridge models, enhancing I2I performance with significant improvements in image quality. The A-Bridge framework \cite{A_bridge} builds on a similar concept with a broader range of generalized Brownian bridges and demonstrates strong performance across various I2I tasks.

\section{Deterministic Dual-approx Bridge}

In this section, we present Dual-approx Bridge in details. The Brownian bridge models \cite{BBDM, A_bridge} have achieved high image quality and diversity in I2I translation. The structure of a Brownian bridge allows it to successfully map between paired probability distributions while generating high quality images of substantial diversity. On the other hand, when it comes to the deterministic I2I translation tasks, diversity is not preferred. In this direction, Dual-approx Bridge delivers deterministic I2I translation with minimal diversity. The key difference between our model and existing diffusion models is that in the generative process, existing models employ one neural network-based approximator which is trained in the forward diffusive process, while we make use of a pair of approximators trained in the forward and reverse diffusive processes separately, shown in Figure \ref{fig:architecture}. 

\subsection{Score-Based SDE}
The score-based Brownian bridge model is prescribed by the forward and reverse SDEs. The forward SDE reads
\begin{equation}\label{eq1}
dX = -\frac{X-Y}{1-t}dt + g(t)\ dW,\ \ t\in (0,1)
\end{equation}
where $W(\cdot)$ is a Brownian motion (i.e., a Wiener process), and $g(t)$ is the variance of the white noise. Then $X(1) = Y$. Let $G(t) = (1-t)^2 \int^t_0 \frac{g^2(s)}{(1-s)^2}\ ds$. It holds that
\begin{align}
\begin{split}
X_t &= (1-t)X_0 + tY + (1-t) \int^t_0 \frac{g(s)}{1-s} \ dW(s)\\
&\sim \mathcal{N}((1-t)X_0+tY, G(t)\ I).
\end{split}
\end{align}
That is
\begin{equation}\label{eq2.3}
X_t = (1-t)X_0 + tY +\sqrt{G(t)}\epsilon,\ \ \epsilon\sim \mathcal{N}(0,I).
\end{equation}
The reverse generative SDE  \cite{ANDERSON1982313} is given by
\begin{equation}\label{eq4}
%\begin{split}
dX = -\left(\frac{X-Y}{1-t} + g^2(t)\nabla \log p(X,t)\right)\ dt + g(t)\ d\overline{W}%\\ 
%&\ \ \ \ \ \ t\in (0,1)
%\end{split}
\end{equation}
where $\overline{W}(\cdot)$ is a reverse time Brownian motion, and the \textit{score function} is given by 
\begin{equation}\label{eq5}
\nabla \log p(X,t) = - \frac{X - (1-t)X_0 - tY}{G(t)}.
\end{equation}
Hence, in the reverse transitional formula where $\Delta t$ is the time increment
\begin{align}
\begin{split}
X_{t-1} &= X_t + \left(\frac{X_t-Y}{1-t} + g^2(t)\nabla \log p(X_t,t)\right)\ \Delta t\\ &\ \ \ \ - g(t)\left(\overline{W}(t) - \overline{W}(t-\Delta t)\right)
\end{split}
\end{align}
we need to estimate two quantities, $X_0$ and $\Delta\overline{W} = \overline{W}(t) - \overline{W}(t-\Delta t) = \sqrt{\Delta t}z$, at each time step. In the SDE model, $z$ is just a random variable of the standard normal distribution. This causes the diversity in the output $\hat{X}_0$. To make the sampling deterministic, we use a neural network $\epsilon_{\theta}$ to estimate $X_0$ in the forward diffusion process and a second neural network $Z_{\phi}(X_t,t)$ to estimate $z$ in the reverse diffusion process. That is, we train a pair of dual approximators $\epsilon_{\theta}$ and $Z_{\phi}$. Note that our goal here is to generate a unique output from a given input, rather than translating probability distributions. So, we need two neural networks to approximate two different samples for the forward and reverse steps from the same white noise distribution. A single approximator cannot serve this purpose.

\subsection{Training of Neural Network Approximators}
For simplicity, we take the variance of the white noise $g(t) = 1$. Then $G(t) = t(1-t)$.

\noindent\textbf{Forward Training Objective:}
We design a neural network $\epsilon_{\theta}$ to minimize, for $t = 1,\hdots,T$, $\epsilon\sim \mathcal{N}(0,I)$,
\begin{align}
\begin{split}
&\|X_t - X_0 - \epsilon_{\theta}(X_t,t)\| =\\ 
&\|t(Y-X_0) + \sqrt{t(1-t)}\epsilon - \epsilon_{\theta}(X_t,t)\|.
\end{split}
\end{align}

Here $X_t$ is defined by Equation \ref{eq2.3} with $G(t) = t(1-t)$. This is the same training process as that of the Brownian bridges\cite{A_bridge}. That is, we do the back propagation of
$$\|t(Y-X_0) + \sqrt{t(1-t)}\epsilon - \epsilon_{\theta}(X_t,t)\|,$$ where $t\sim \textrm{Uniform}(1,\hdots,T)$ and $\epsilon\sim\mathcal{N}(0,I)$. Essentially, the neural network $\epsilon_{\theta}$ is used to estimate $X_0$ given $X_t$ at the time step $t$. Note that an optional conditional network design $\epsilon_{\theta}(X_t,t, Y)$ can also be used, which may improve the stability and efficiency of model training. 

\noindent\textbf{Reverse Training Objective:} Independently of the forward training, we design a neural network $Z_{\phi}(X_t,t)$ to minimize, for $t = 1,\hdots,T$,
\begin{equation}\label{eq3.2}
\|z - Z_{\phi}(X_t,t)\|,\ \ z\sim \mathcal{N}(0,I),
\end{equation}
where $X_t$ is defined by $X_t = (1-t)X_0 + tY + \sqrt{t(1-t)}z$. That is, we do the back propagation of
$$\|z - Z_{\phi}(X_t,t)\|,$$ where $t\sim \textrm{Uniform}(0,\hdots,T-1)$ and $z\sim\mathcal{N}(0,I)$.

%We notice that the reverse training does not do for $t=T$, i.e., $Z_{\phi}(X_T,T)$ is not defined.

\subsection{Sampling}
We discretize Equation \ref{eq4} with $g(t) = 1$ to get the iterative generative formula:
\begin{equation}\label{eq4.1}
X_{t-1} = \left(1-\frac{\Delta t}{t}\right)X_t + \frac{\Delta t}{t}X_0 - \left[\overline{W}(t) - \overline{W}(t-\Delta t)\right]
\end{equation}

We employ the two neural networks in this way: $X_0 \approx X_t - \epsilon_{\theta}(X_t,t)$, while $\overline{W}(t) - \overline{W}(t-\Delta t)$ must be approximated with the use of the second neural network $Z_{\phi}$.

We make an observation here: $\overline{W}(t) - \overline{W}(t-\Delta t)$ and $\overline{W}(t+\Delta t) - \overline{W}(t)$ are independent, equivalent random variables. Let $g(t) = 1$ in (\ref{eq4}) and (\ref{eq5}) to get $X_t = (1-t)X_0 + tY + t\int^t_1\frac{d\overline{W}}{s}$. Then we get
\begin{align}
\begin{split}
&\overline{W}(t) - \overline{W}(t-\Delta t) = \overline{W}(t+\Delta t) - \overline{W}(t) \approx\\ &U(X_{t+1},t+1) - \left(1 + \frac{\Delta t}{t}\right)U(X_t,t) + \frac{\Delta t}{t}(X_0-Y),
\end{split}
\end{align}
where 
\begin{equation}
U(X_t,t) = X_t - Y = (1-t)(X_0-Y) + t\int^t_1\frac{d\overline{W}}{s}.
\end{equation}
Hence, by plugging the above estimate into Equation \ref{eq4.1}, we get the final iterative generative formula
\begin{align}
\label{eq4.3}
\begin{split}
X_{t-1} &= \left(1-\frac{\Delta t}{t}\right)X_t + \frac{\Delta t}{t}Y - U(X_{t+1},t+1)\\ 
&+ \left(1 + \frac{\Delta t}{t}\right)U(X_t,t).
\end{split}
\end{align}

In the formula (\ref{eq4.3}), replace $t$ by $m_t = \frac{t}{T}$ and $\Delta t$ by $\frac{1}{T}$ if we apply the uniform sampling schedule with $T$ time steps. Also, we estimate $X_0$ by
\begin{equation}
\hat{X}_0 = X_t - \epsilon_{\theta}(X_t,t).
\end{equation}
Then for $t = T-1,T-2,\hdots,2$, we get
\begin{align}
\begin{split}
X_{t-1} &= \left(1-\frac{1}{t}\right)X_t + \frac{1}{t}Y - U(X_{t+1},t+1)\\ &+ \left(1+\frac{1}{t}\right)U(X_t,t).
\end{split}
\end{align}

\begin{algorithm}
\caption{Forward Approximator Training Algorithm}
\label{alg:train:forward}
\begin{algorithmic}[1]
\STATE \textbf{Input} Paired Dataset $D_{x,y}$, Forward Network $\epsilon_{\theta}$, Total Time Steps $T$
    \REPEAT
        \STATE $x,y \sim D_{x,y}$
        \STATE $t \sim \text{Uniform}(1, \ldots, T)$
        \STATE $\epsilon \sim \mathcal{N}(0, I)$
        \STATE $x_t = (1-\frac{t}{T} )x_0 + \frac{t}{T}y + B(t)\epsilon$
        \STATE $\nabla_\theta \left\lVert x_t - x_0 - \epsilon_{\theta}(x_t, t) \right\rVert$
    \UNTIL{converged}
    \STATE \textbf{end function}
\end{algorithmic}
\end{algorithm}

\begin{algorithm}
\caption{Reverse Approximator Training Algorithm}
\label{alg:train:backward}

\begin{algorithmic}[1]
\STATE \textbf{Input} Paired Dataset $D_{x,y}$, Reverse Network $Z_{\phi}$, Total Time Steps $T$
    \REPEAT
        \STATE $x,y \sim D_{x,y}$
        \STATE $t \sim \text{Uniform}(1, \ldots, T)$
        \STATE $z \sim \mathcal{N}(0, I)$
        \STATE $x_t = (1-\frac{t}{T} )x_0 + \frac{t}{T}y + B(t)z$
        \STATE $\nabla_\phi \left\lVert z - Z_{\phi}(x_t, t) \right\rVert$
    \UNTIL{converged}
    \STATE \textbf{end function}
\end{algorithmic}
\end{algorithm}

This iterative formula does not apply when $t = T$ as $X_{T+1}$ does not exist. So we define $X_{T-1}$ as we would in the SDE model
\begin{equation}
X_{T-1} = Y - \frac{1}{T}\epsilon_{\theta}(Y,T) - \frac{1}{\sqrt{T}}z,\ \ z\sim \mathcal{N}(0,I).
\end{equation}

Finally, we define
\begin{equation}
\hat{X}_0 = X_1 - \epsilon_{\theta}(X_1,1)
\end{equation}
as the final output of the algorithm. 

The only randomness introduced in the entire sampling process is in $X_{T-1}$, which should be negligible, as the coefficient  $\frac{1}{\sqrt{T}}$ of $z$ is fairly small compared to that of $Y$. On the contrary, SDE models introduce randomness at every sampling step and hence cause substantial diversity to the output.

\subsection{Algorithm}
In a given paired dataset $D_{x,y}$, $(x,y)$ denotes a paired images from the set. Define $B(t) = G\left(\frac{t}{T}\right) = \frac{1}{T}\sqrt{t(T-t)}$, and we have the training and sampling algorithms shown in Algorithms 1 to 3, respectively.

% \begin{algorithm}
%     \caption{Sampling Algorithm}
%     \label{alg:sampling}
%     \renewcommand{\algorithmicrequire}{\textbf{Input:}}
%     \renewcommand{\algorithmicensure}{\textbf{Output:}}
    
%     \begin{algorithmic}[1]
%         \REQUIRE $y$, $\epsilon_{\theta}$, $Z_{\phi}$ %%input
%         \ENSURE Deterministic Result of $x_{0}$   %%output
%         \STATE $x_{t}=y$
%         \FOR{ $t = T,...,1$}  
%             \IF {$ t > 1$}
%                 \STATE $\hat{x}_0 = x_t - \epsilon_{\theta}(x_t,t, \textcolor{red}{y})$
%                 \IF {$t == T$}
%                     \STATE $\hat{\epsilon} \sim \mathcal{N}(0, I)$
%                 \ELSE
%                     \STATE $\hat{\epsilon} = Z_{\phi}(x_{t},t)$
%                 \ENDIF
%                 \STATE$m_t = \frac{t}{T}$
%                 \STATE$m_{t-1} = \frac{t-1}{T}$
%                 \STATE$c_{noise} = \frac{1}{\sqrt{T}} \left( -\sqrt{t \left(1 - m_t \right)} 
% + \frac{t}{t-1} \cdot \sqrt{(t-1) \left(1 - m_{t-1} \right)} \right)$
%                 \STATE $x_{t-1}=x_t - \frac{1}{t}\epsilon_{\theta}(x_t,y,t) + c_{noise} \hat{z}$
%             \ELSE
%                 \STATE $x_{0} = \hat{x}_0 $
%             \ENDIF
            
%         \ENDFOR
%         \RETURN $x_{0}$
%     \end{algorithmic}
% \end{algorithm}
\begin{algorithm}
    \caption{Sampling Algorithm (Refactored)}
    \label{alg:sampling}
    \renewcommand{\algorithmicrequire}{\textbf{Input:}}
    \renewcommand{\algorithmicensure}{\textbf{Output:}}
    
    \begin{algorithmic}[1]
        \REQUIRE $y$, $\epsilon_{\theta}$, $Z_{\phi}$
        \ENSURE $x_{0}$  
        \STATE $x_{t} = y$
        \FOR{ $t = T$ to $1$}  
            \STATE $\hat{x}_0 = x_t - \epsilon_{\theta}(x_t, t)$
            \STATE $\hat{z} \sim \mathcal{N}(0, I)$ if $t == T$ else $\hat{z} = Z_{\phi}(x_{t}, t)$
            \STATE $\hat{x}_{t} = (1 - \frac{t}{T})\hat{x}_{0} + \frac{t}{T} y + B(t)\hat{z}$
            \STATE $\hat{x}_{t-1} = (1 - \frac{t-1}{T})\hat{x}_{0} + \frac{t-1}{T} y + B(t - 1)\hat{z}$
            \STATE $U_{t+1} = \hat{x}_{t} - y, U_{t} = \hat{x}_{t-1} - y$
            \STATE $x_{t-1}= (1 - \frac{1}{t})x_{t} + \frac{1}{t-1}y - \frac{1}{t(t-1)}\hat{x}_{0} - U_{t+1} + \frac{t}{t-1}U_{t}$
        \ENDFOR
        \STATE $x_{0} =\hat{x}_0 $
        \RETURN $x_{0}$
    \end{algorithmic}
\end{algorithm}

% \begin{algorithm}
%     \caption{Sampling Algorithm}
%     \label{alg:sampling}
%     \renewcommand{\algorithmicrequire}{\textbf{Input:}}
%     \renewcommand{\algorithmicensure}{\textbf{Output:}}
    
%     \begin{algorithmic}[1]
%         \REQUIRE $y$, $\epsilon_{\theta}$, $Z_{\phi}$ %%input
%         \ENSURE Deterministic Result of $x_{0}$   %%output
%         \STATE $x_{t}=y$
%         \FOR{ $t = T,...,1$}  
%             \IF {$ t > 1$}
%                 \STATE $\hat{x}_0 = x_t - \epsilon_{\theta}(x_t,t)$
%                 \IF {$t == T$}
%                     \STATE $\hat{z} \sim \mathcal{N}(0, I)$
%                 \ELSE
%                     \STATE $\hat{z} = Z_{\phi}(x_{t},t)$
%                 \ENDIF
                
%                 \STATE $\hat{x}_{t} = (1-\frac{t}{T} )\hat{x}_{0} + \frac{t}{T}y + B(t)\hat{z}$
%                 \STATE $\hat{x}_{t-1} = (1-\frac{t-1}{T} )\hat{x}_{0} + \frac{t-1}{T}y + B(t - 1)\hat{z}$
%                 \STATE $U_{t+1} = \hat{x}_{t} - y$
%                 \STATE $U_{t} = \hat{x}_{t-1} - y$
%                 \STATE $x_{t-1}= (1 - \frac{1}{t})x_{t} + \frac{1}{t-1}y - \frac{1}{t(t-1)}\hat{x}_{0} - U_{t+1} + \frac{t}{t-1}U_{t}$
%             \ELSE
%                 \STATE $x_{0} =\hat{x}_0 $
%             \ENDIF
            
%         \ENDFOR
%         \RETURN $x_{0}$
%     \end{algorithmic}
% \end{algorithm}

\section{Experiments}
In this section, we verify the capability of Dual-approx Bridge, addressing the following questions:

\begin{itemize}
    \item How do varying the number of sampling steps impact its performance in terms of image quality and faithfulness to GT? Furthermore, how does the sampling variance of our model compare to that of SDE-based samplers in terms of faithfulness to GT?
    \item How does Dual-approx Bridge perform in terms of faithfulness to GT compared to other models in image-to-image translation tasks that demand high faithfulness and deterministic result?
    \item How does Dual-approx Bridge perform in general image-to-image translation tasks, which prioritize image quality, when compared to other models including GANs, Diffusion models, and Bridge models?
\end{itemize}
To address these questions, we used datasets with the following resolutions: Cityscapes \cite{cordts2016cityscapes} (256×256), Edges2Handbags \cite{zhu2016edges2handbags} (64×64), DF2K \cite{timofte2017ntire} (256×256), Urban100 \cite{huang2015single} (256×256 patches), and BSD100 \cite{martin2001database} (256×256 patches). For evaluation, we used FID \cite{heusel2017gans} to assess image realism (lower is better), LPIPS \cite{zhang2018perceptual} for perceptual similarity (lower is better), PSNR for pixel-level similarity (higher is better) and SSIM \cite{wang2004image} for structural similarity (closer to 1 is better). Our super-resolution (SR) experiments followed the experimental design of \cite{chen2023solving}. All of our experiments were implemented using PyTorch \cite{pytorch} and were run on a machine with two NVIDIA RTX A6000 GPUs.

\subsection{Ablation Study}

\subsubsection{Impact of Sampling Steps}

% We first evaluate the impact of sampling steps on the performance of our model on the Cityscapes dataset, which prioritizes both image quality and fidelity. We assessed performance across two metrics: FID + LPIPS (image quality) and PSNR + SSIM (fidelity). We evaluated this across four different step counts: 1,000, 200, 10, and 3.
We first evaluate the impact of sampling steps on our model's performance in the label-to-photo translation task on the Cityscapes dataset. Since the forward approximator in our model can predict $x_0$ given $x_t$, sampling can be stopped at any time, allowing us to test sampling performance with a different number of steps. Table \ref{tab:sampling:steps} presents how the varying sampling steps affect both image quality and faithfulness. It demonstrates that 1,000 sampling steps yield satisfactory image quality, with FID reaching a minimum of 42.70 and LPIPS dropping to 0.442. However, this comes at the cost of reduced faithfulness when compared to results obtained using fewer sampling steps, as evidenced by the decline in PSNR and SSIM. Notably, with only three sampling steps, our model achieves its peak in terms of faithfulness, attaining 16.99 PSNR and 57.4\% SSIM. This demonstrates the model’s adaptability to meet the diverse requirements of various applications.

Additionally, we assessed the variance of the output fidelity in five independent sampling trials given the same input. Clearly, even at a higher sampling step of 1,000, the standard deviation of PSNR and SSIM remained low (0.02 and 0.07\%, respectively), and it approaches zero with three sampling steps. This robustness underscores the superiority of our model in consistently producing high-quality and faithful results with minimal variability. Consequently, we employed \textit{three sampling steps} for better robustness and higher efficiency in our subsequent experiments, unless otherwise specified. Our experiments show that three iterations perform well across these metrics,  though further research is needed to determine the optimal number of sampling steps for different tasks.

\begin{table}
    \centering
    \small
    \resizebox{0.9\columnwidth}{!}{
    \begin{tabular}{c | c c c c}
        \hline
        \textbf{Steps} & \textbf{FID} $\downarrow$ & \textbf{LPIPS} $\downarrow$  & \textbf{PSNR} $\uparrow$ & \textbf{SSIM \% } $\uparrow$ \\
        \hline
        \textbf{1,000} & \textbf{42.70} & \textbf{0.442} & 15.10  $\pm 0.02$ & (43.2  $\pm 0.07$) \% \\
        \textbf{200} & 48.70 & 0.492 & 15.70  $\pm 0.02$ & (53.2  $\pm 0.06$) \% \\
        \textbf{10} & 57.43 & 0.530 & 15.62 $\pm 0.02$ & (55.3 $\pm 0.06$) \%\\
        \textbf{3} & 56.07 & 0.523 & \textbf{16.99 $\pm 0.01$} & (\textbf{57.4 $\pm 
  0.00$}) \%\\
        \hline
    \end{tabular}
    }
    \caption{ Quantitative comparison with different sampling steps on the Cityscapes dataset. \textbf{Bold} values represent the best performance. Standard Deviation is obtained by five sampling trials for the same input. }
    \label{tab:sampling:steps}
\end{table}

\subsubsection{Faithfulness Evaluation}
\label{subsec:Faithfulness}
\begin{table}
    \centering
    \resizebox{0.9\columnwidth}{!}{%
    \begin{tabular}{c | c c}
        \hline
        \textbf{Method}  & \textbf{PSNR} $\uparrow$ & \textbf{SSIM \%} $\uparrow$ \\      
        \hline
        Brown Bridge Model (SDE)& 14.28  $\pm 0.21$  & (49.6   $\pm 0.63$) \%\\
        Brown Bridge Model (PF-ODE) & 15.23  $\pm 0.00$ & (51.2  $\pm 0.00$) \%\\
        \hline
        \textbf{Ours} & \textbf{16.99} $\pm 0.01$ & (\textbf{57.4} $\pm 0.00$)\%\\
        \hline
    \end{tabular}%
    }
    \caption{Quantitative comparison with Brownian bridges using SDE and PF-ODE-based samplers on Cityscape. \textbf{Bold} values represent the best performance. Standard deviation is obtained by five sampling trials for the same input. }
    \label{tab:cityscape:var}
\end{table}
\begin{figure}[h]
    \centering
    \includegraphics[width=\columnwidth]{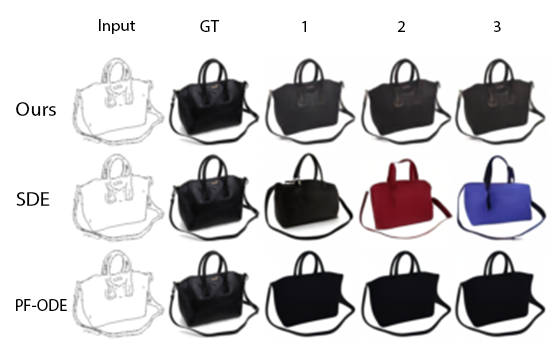}
    \caption{Qualitative comparison with Brownian bridges using SDE-based and PF-ODE-based samplers on Edges2Handbags. Three sampling trials are conducted by each model given the same input. Standard deviation among sampling results by our model, SDE-based and PF-ODE-based sampler are 0.0001, 0.1030 and 0, respectively.}
    \label{fig:var}
\end{figure}

We then compare the faithfulness of Dual-approx Bridge to other Brownian bridge models with SDE-based \cite{A_bridge,BBDM} and PF-ODE-based samplers on the Cityscapes dataset using PSNR and SSIM metrics. Both the SDE-based and PF-ODE-based samplers use 1,000 sampling steps. The Brownian bridge model, as the leading I2I translation method, is an ideal benchmark for us due to the same bridge structure.

As shown in Table \ref{tab:cityscape:var}, our Dual-approx Bridge clearly outperforms SDE-based and PF-ODE-based samplers. The PSNR value increased by approximately 12\% compared to the second-best, the PF-ODE-based sampler, while the SSIM value showed an overall improvement of 12\%, rising from 51.2\% to 57.4\%. Notably, the metrics of the Dual-approx Bridge exhibited minimal variance, whereas SDE-based sampler exhibited substantial variance in PSNR and SSIM due to the presence of white noise in every step of its sampling.

Figure \ref{fig:var} shows three sample images by each model with the same input in the Edges2Handbags dataset. The images generated by SDE-based sampler are rich in detail but distinct from each other. The PF-ODE-based sampler produces consistent results but lacks details, resulting in a blurred appearance. As a comparison, Dual-approx Bridge achieves SDE-level quality with high and consistent faithfulness. 

% demonstrates that Dual-approx Bridge maintains the stability of PF-ODE models while enhancing detail preservation, capturing the benefits of SDE-level detail without sacrificing faithfulness. It effectively overcomes the limitations of both approaches, providing the high detail of SDE models without variance issues and the faithfulness of PF-ODE models without compromising image quality.

\subsection{Deterministic Image-to-Image Translation}
We selected the label-to-photo task in the Cityscapes dataset for deterministic image generation, where the goal is to generate realistic images from segmentation maps. Faithfulness is essential in this task, as it ensures the generated images accurately reflect the structure and content of the input labels, preserving the integrity of labeled features. We also included image super-resolution as a deterministic image-to-image translation task, which also requires both high quality and faithfulness.
\begin{table}
    \centering
    \resizebox{0.8\columnwidth}{!}{%
    \begin{tabular}{c | c c c}
        \hline
        \textbf{Method} & \textbf{FID} $\downarrow$  & \textbf{PSNR} $\uparrow$ & \textbf{SSIM \%} $\uparrow$ \\
        
        \hline
        Pix2pix & 101.04 & 12.86   & 28.63  \%\\
        CycleGAN & 75.37  & 14.53   & 34.16  \%\\
        UNIT & 82.74  & 14.61   & 35.41  \%\\
        \hline
        \textbf{Ours} & \textbf{48.70} & \textbf{15.70} & \textbf{53.26} \%\\
        \hline
    \end{tabular}
    }
    \caption{Quantitative comparison with deterministic models on Cityscape. Results of our model are obtained using 200 sampling steps. \textbf{Bold} values represent the best performance.}
    \label{tab:cityscape:faithfulness}
    % \vspace{-0.1in}
\end{table}

Given the inherent randomness in diffusion models, we primarily chose GAN-based models as baselines for deterministic I2I tasks. We selected Pix2Pix \cite{pix2pix}, CycleGAN \cite{cycleGAN}, and UNIT \cite{UNIT} for comparison on the Cityscapes dataset. For image super-resolution, we used ESRGAN \cite{ESRGAN}, RankSRGAN \cite{RankSRGAN}, and SRDiff \cite{SRDiff} as baselines and followed the experimental configuration in \cite{ma2023solving} to train our model on DF2K and test it on Urban100 and BSD100.

\begin{figure}
    \centering
    % \hspace{-0.1\columnwidth}
    \includegraphics[width=0.9\columnwidth]{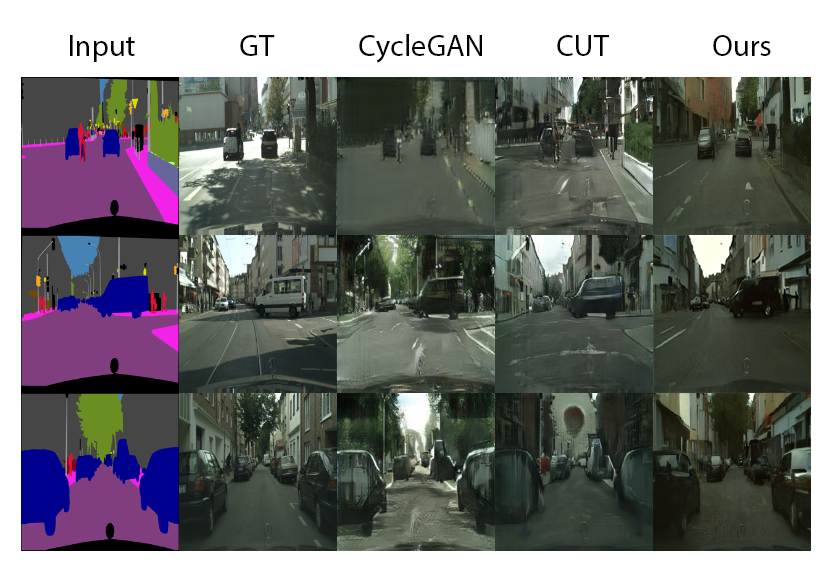}
    % \vspace{-0.2in}
    \caption{Qualitative comparison between Dual-approx Bridge and SOTA methods on the Cityscapes dataset.}
    \label{fig:cityscape}
    % \vspace{-0.2in}
\end{figure}

% \begin{figure}
%     \centering
%     \includegraphics[width=\columnwidth]{imgs/sr.png}
%     \caption{Qualitative comparison on Urban100 and BSD100 datasets. }
%     \label{fig:sr}
% \end{figure}

\begin{figure}
    \centering
    \includegraphics[width=\columnwidth]{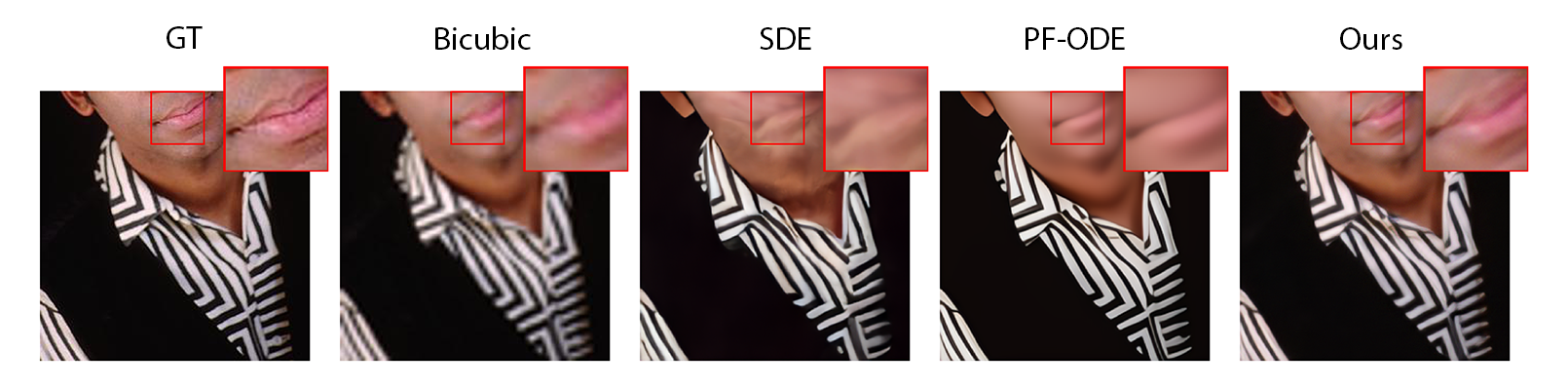}
    \caption{Qualitative comparison on BSD100 datasets. }
    \label{fig:sr}
\end{figure}

\begin{figure}
    \centering
\includegraphics[width=0.95 \columnwidth]{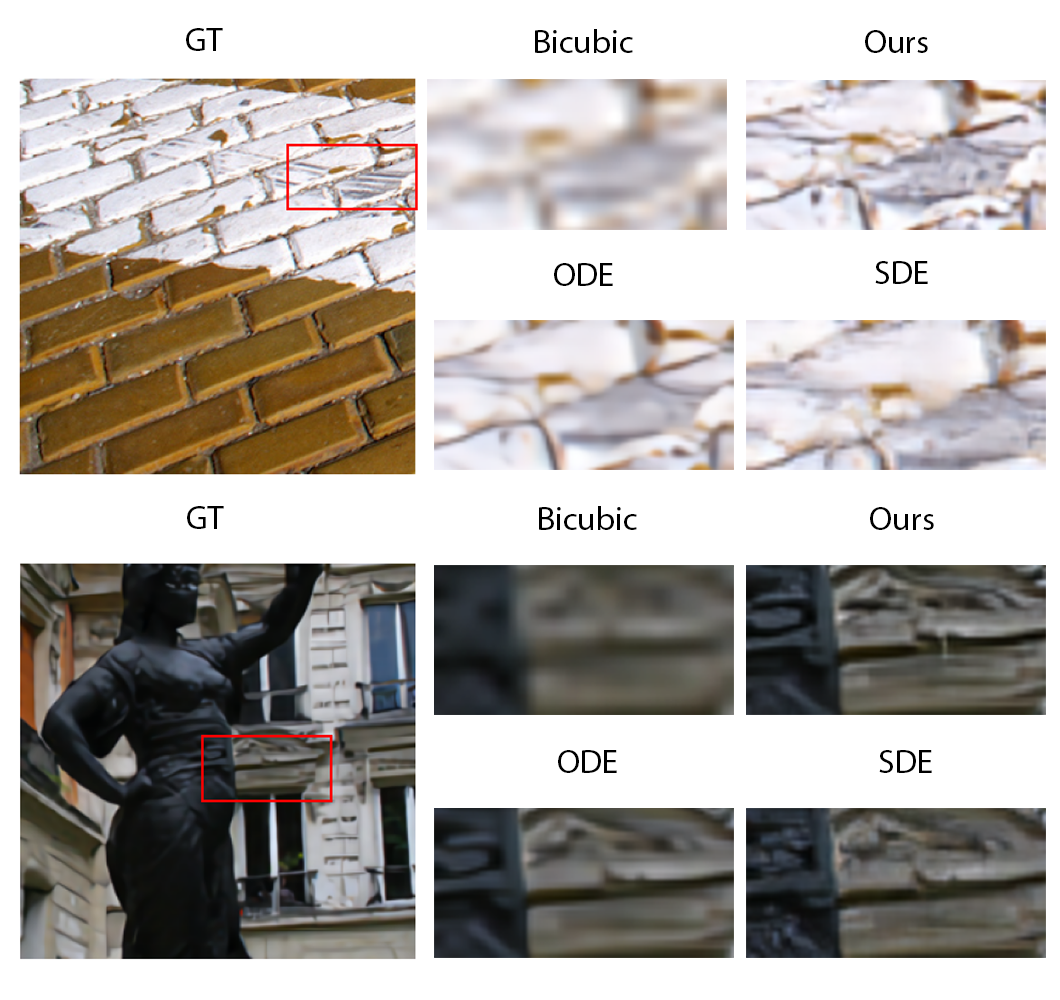}
    \caption{Qualitative comparison on Uban100 datasets. }
    \label{fig:sr_urben}
\end{figure}

Table \ref{tab:cityscape:faithfulness} compares our model against baseline models on the Cityscapes dataset. The Dual-approx Bridge model demonstrates superior performance across all metrics compared to the other models. In terms of image quality, our model achieves a significant improvement in FID, with a score of 48.70 compared to the second-best score of 75.37, representing a substantial 35\% performance enhancement. Furthermore, our model exhibits notable gains in PSNR and SSIM, with PSNR increased from 14.61 to 15.70 and SSIM improved by approximately 50\% (from 35.41\% to 53.26\%). Qualitative comparisons of various methods are presented in Figure \ref{fig:cityscape}.

% \begin{table}
%     \centering
%     \small
%     \resizebox{0.45\textwidth}{!}{%
%     \begin{tabular}{c | c c c c c c}
%         \hline
%         \multirow{2}{*}{\textbf{Model}} & \multicolumn{2}{c}{\textbf{Urban100}} & \multicolumn{2}{c}{\textbf{BSD100}} \\
%          & \textbf{LPIPS} $\downarrow$ & \textbf{PSNR} $\uparrow$ & \textbf{LPIPS} $\downarrow$ & \textbf{PSNR} $\uparrow$ \\
%         \hline
%         Bicubic & 0.4826 & 21.75 & 0.5282 & 24.32  \\
%         \hline
%         ESRGAN & 0.1226 & 23.04 & 0.1579 & 23.65  \\
%         RankSRGAN & 0.1403 & 23.16 & 0.1714 & 23.80  \\
%         SRDiff & 0.1391 & \textbf{23.88} & 0.2046 & 24.17 \\
%         \hline
%         \textbf{Ours} & \textbf{0.1207} & 23.23 & \textbf{0.1267} & \textbf{26.56}  \\
%         \hline
%     \end{tabular}%
%     }
%     \caption{Quantitative comparisons on Urban100 and BSD100}
%     \label{tab:super-reso}
%     % \vspace{-0.25in}
% \end{table}

\begin{table}
    \centering
    \small
    \resizebox{0.9\columnwidth}{!}{%
    \begin{tabular}{c | c c c c c c}
        \hline
        \multirow{2}{*}{\textbf{Model}} & \multicolumn{2}{c}{\textbf{BSD100}} & \multicolumn{2}{c}{\textbf{Urban100}} \\
         & \textbf{LPIPS} $\downarrow$ & \textbf{PSNR} $\uparrow$ & \textbf{LPIPS} $\downarrow$ & \textbf{PSNR} $\uparrow$ \\
        \hline
        Bicubic & 0.5282 & 24.32 & 0.4826 & 21.75  \\
        \hline
        ESRGAN & 0.1579 & 23.65 & 0.1226 & 23.04  \\
        RankSRGAN & 0.1714 & 23.80 & 0.1403 & 23.16  \\
        SRDiff & 0.2046 & 24.17 & 0.1391 & \textbf{23.88} \\
        \hline
        \textbf{Ours} & \textbf{0.1267} & \textbf{26.56} & \textbf{0.1207} & 23.23  \\
        \hline
    \end{tabular}%
    }
    \caption{Quantitative comparisons on BSD100 and Urban100. \textbf{Bold} values represent the best performance.}
    \label{tab:super-reso}

\end{table}

\begin{figure*}
    \centering
    \includegraphics[width=0.9\textwidth]{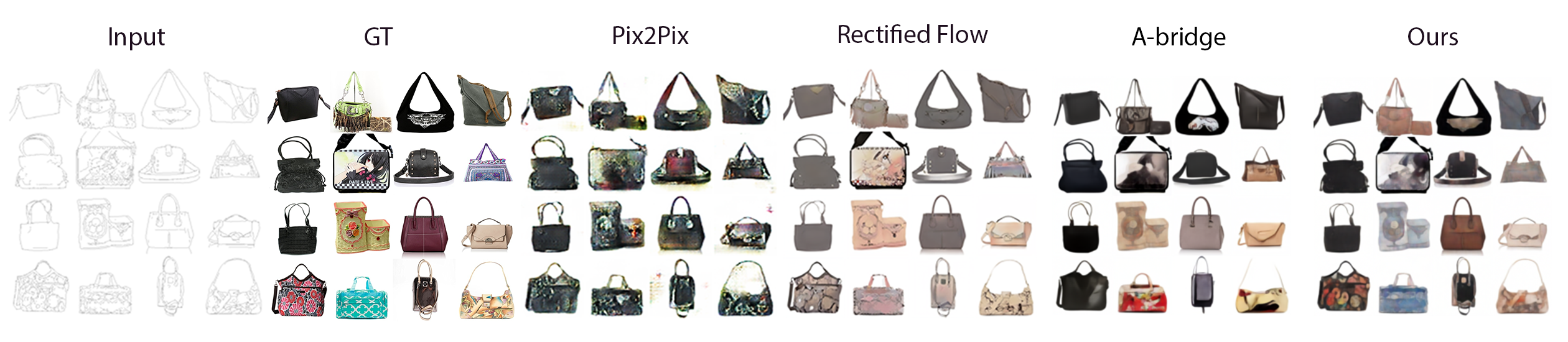}
    \vspace{-0.1in}
    \caption{Qualitative comparison between Dual-approx Bridge and SOTA methods on the Edges2handbags dataset.}
    \label{fig:edges2handbags_compare}
    \vspace{-0.1in}
\end{figure*}

Table \ref{tab:super-reso} compares our model against baselines on super-resolution tasks. Our model performed the best on the BSD100 dataset with LPIPS at 0.1267 and PSNR at 26.56. It also achieves the highest LPIPS score of 0.1207 and the second-highest PSNR score of 23.23 on the Urban100 dataset. While SRDiff achieved the best PSNR on Urban100, its diffusion process encounters challenges in consistently producing stable outputs due to the inherent randomness in its SDE-based sampler (as shown in Table \ref{tab:cityscape:var}). Moreover, our model accomplishes this performance with significantly fewer sampling steps (3 used in our model vs. 1,000 used in SRDiff).

Figure \ref{fig:sr} compares the visual quality of various methods for image super-resolution. The top-right of each image shows a zoomed-in view of the red-boxed region for a detailed comparison. Due to inherent randomness, SDEs often introduce unrealistic artifacts, as seen in the example. The PF-ODE-based sampler results exhibit excessive smoothness and the loss of details, as evident in the zoomed regions for both examples. Figure \ref{fig:sr_urben} further illustrates that our model can achieve detail preservation approaching or even surpassing that of SDE-based methods. Conversely, the Dual-approx Bridge method demonstrates a remarkable ability to preserve details while maintaining high faithfulness with GT.

\begin{table}
    \centering
    \resizebox{0.7\columnwidth}{!}{
    \begin{tabular}{w{c}{1in} | w{c}{1in}}
        \hline
        \textbf{Method} & \textbf{FID} $\downarrow$\\
        \hline
        MUNIT & 91.4  \\
        DRIT & 155.3  \\
        Distance & 85.8  \\
        SelfDistance & 78.8   \\
        CGGAN & 105.2 \\
        CUT & 56.4 \\
        FastCUT & 68.8  \\
        DCLGAN & 49.4  \\
        SimDCL  & 51.3  \\
        \hline
        \textbf{Ours} & \textbf{48.7} \\
        \hline
    \end{tabular}
    }
    \caption{Quantitative comparison of image quality on Cityscapes (label to photo). \textbf{Bold} values represent the best performance.}
    \label{tab:cityscape}
    \vspace{-0.1in}
\end{table}

% \begin{figure}
%     \centering
%     % \hspace{-0.1\textwidth}
%     \includegraphics[width=\columnwidth]{imgs/edges2handbags_compare_2_rows.png}
%     \caption{The qualitative comparison between our Dual-approx Bridge model and state-of-the-art methods on Edges2handbags dataset.}
%     \label{fig:edges2handbags_compare}
%     \vspace{-0.2in}
% \end{figure}

\begin{figure}
    \centering
    \includegraphics[width=0.9\columnwidth]{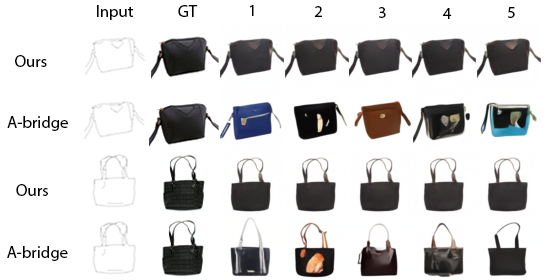}
    \caption{Quantitative comparison between Dual-approx Bridge and A-bridge (with SDE-based sampler). Each model are sampled five times given the same input. Sampling standard deviation of the first example (Ours: 0.0084, A-bridge: 0.0903) and the second example (Ours: 0.0054, SDE: 0.0672).}
    \label{fig:edges2handbags_additional_var}
    \vspace{-0.2in}
\end{figure}

\subsection{General Image-to-Image Translation}
Finally, we evaluate Dual-approx Bridge on general image-to-image translation tasks. We assess our model’s image quality on the Edges2Handbags and Cityscapes datasets. For Edges2Handbags, we conduct comparisons with Pix2Pix \cite{pix2pix}, SDEdit \cite{meng2021sdedit}, DDIB \cite{su2022dual}, Rectified Flow \cite{liu2022flow}, I2SB \cite{liu20232}, DDBM \cite{DDBM}, and A-bridge \cite{A_bridge}. 
For the Cityscapes dataset, we compared our model with MUNIT \cite{huang2018multimodal}, DRIT \cite{lee2018diverse}, DistanceGAN \cite{benaim2017one}, Self-DistanceGAN \cite{benaim2017one}, CGGAN \cite{zhou2020cggan}, CUT \cite{park2020contrastive}, FastCUT \cite{park2020contrastive}, DCLGAN \cite{han2021dual}, and SimDCL \cite{han2021dual}. 
% Our empirical comparison encompasses a diverse set of approaches, including GAN-, diffusion-, bridge-, and flow-based methods, representing key paradigms in I2I translations.

On the Cityscapes dataset, our model outperforms all GAN-based models, achieving the top position and surpassing all competitors (Table \ref{tab:cityscape}). As shown in Table \ref{tab:Edges2Handbags}, our model achieves 1.36 FID score on the Edges2Handbags dataset, securing the second place. This result demonstrates that, despite our primary focus on enhancing faithfulness, our model still delivers highly competitive image quality. 

Figure \ref{fig:edges2handbags_compare} provides a qualitative comparison of our model with A-bridge on the Edges2Handbags dataset. Although A-bridge achieves the best FID score, it uses SDE-based samplers. As shown in Figure \ref{fig:edges2handbags_additional_var}, our model demonstrates much higher faithfulness and consistency across repeated sampling with far fewer sampling steps (3 in our model vs. 50 in A-bridge).

\begin{table}
    \centering
    \resizebox{0.7\columnwidth}{!}{
    \begin{tabular}{w{c}{1in} | w{c}{1in}}
        \hline
        \textbf{Method} & \textbf{FID} $\downarrow$\\
        \hline
        Pix2Pix & 74.8  \\
        DDIB &  186.84  \\
        SDEdit & 85.8  \\
        SelfDistance & 26.5   \\
        Rectified Flow & 25.3 \\
        I2SB & 7.43 \\
        DDBM (VP) & 1.83  \\
        A-Bridge & \textbf{1.07 }\\
        \hline
        \textbf{Ours} & 1.36 \\
        \hline
    \end{tabular}
    }
    \caption{Quantitative comparison of image quality on Edges2Handbags. \textbf{Bold} values represent the best performance.}
    \label{tab:Edges2Handbags}
    \vspace{-0.2in}
\end{table}

\section{Conclusion and Future Work}
In this paper, we proposed Dual-approx Bridge, a novel generative model to produce faithful and high quality output for deterministic I2I translations. Our extensive experiments on benchmark datasets demonstrate the consistent and superior performance of Dual-approx Bridge when compared to SOTA I2I models. Our current choice of the SDE follows the Variance Preserving (VP) formulation. Moving forward, we plan to explore the potentials of combining both VP and variance exploding SDEs to further enhance our model's capabilities.

\clearpage
\newpage
{
    \small
    \bibliographystyle{ieeenat_fullname}
    \bibliography{main}
}


\end{document}